\pdfoutput=1

\documentclass[11pt]{article}
\usepackage[utf8]{inputenc}
\usepackage[T1]{fontenc}
\usepackage{CJKutf8}  

\usepackage[]{ACL2023}

\usepackage{times}
\usepackage{latexsym}
\usepackage{booktabs}
\usepackage{caption}
\usepackage[justification=centering]{caption}

\usepackage[T1]{fontenc}

\usepackage[utf8]{inputenc}

\usepackage{microtype}

\usepackage{inconsolata}
\usepackage{graphicx}
\usepackage{hyperref}

\title{ArcMMLU: A Library and Information Science Benchmark for Large Language Models}

\author{Shitou Zhang\textsuperscript{1,2}, Zuchao Li\textsuperscript{3,*}, Xingshen Liu\textsuperscript{2}, Liming Yang\textsuperscript{4}, Ping Wang\textsuperscript{1,2,}\thanks{\quad Corresponding author}\\
        \textsuperscript{1}Key Laboratory of Archival Intelligent Development and Service, NAAC \\
        \textsuperscript{2}School of Information Management, Wuhan University; \\
        \textsuperscript{3}School of Computer Science, Wuhan University \\
        \textsuperscript{4}School of Law, Tsinghua University \\
        \texttt{\{shitouzhang, zcli-charlie, XingshenLiu, wangping\}}@whu.edu.cn\\  
        \texttt{yanglm23}@mails.tsinghua.edu.cn}

\begin{document}
\begin{CJK*}{UTF8}{gbsn}  
\maketitle
\begin{abstract}

In light of the rapidly evolving capabilities of large language models (LLMs), it becomes imperative to develop rigorous domain-specific evaluation benchmarks to accurately assess their capabilities. In response to this need, this paper introduces ArcMMLU, a specialized benchmark tailored for the Library \& Information Science (LIS) domain in Chinese. This benchmark aims to measure the knowledge and reasoning capability of LLMs within four key sub-domains: Archival Science, Data Science, Library Science, and Information Science.
Following the format of MMLU/CMMLU, we collected over 6,000 high-quality questions for the compilation of ArcMMLU. This extensive compilation can reflect the diverse nature of the LIS domain and offer a robust foundation for LLM evaluation.
Our comprehensive evaluation reveals that while most mainstream LLMs achieve an average accuracy rate above 50\% on ArcMMLU, there remains a notable performance gap, suggesting substantial headroom for refinement in LLM capabilities within the LIS domain. Further analysis explores the effectiveness of few-shot examples on model performance and highlights challenging questions where models consistently underperform, providing valuable insights for targeted improvements. ArcMMLU fills a critical gap in LLM evaluations within the Chinese LIS domain and paves the way for future development of LLMs tailored to this specialized area\footnote{The data and code of ArcMMLU are publicly available at \href{https://github.com/stzhang-patrick/ArcMMLU}{https://github.com/stzhang-patrick/ArcMMLU}}.

\end{abstract}

\section{Introduction}

\begin{figure*}[ht!]
\centering
\includegraphics[scale=0.5]{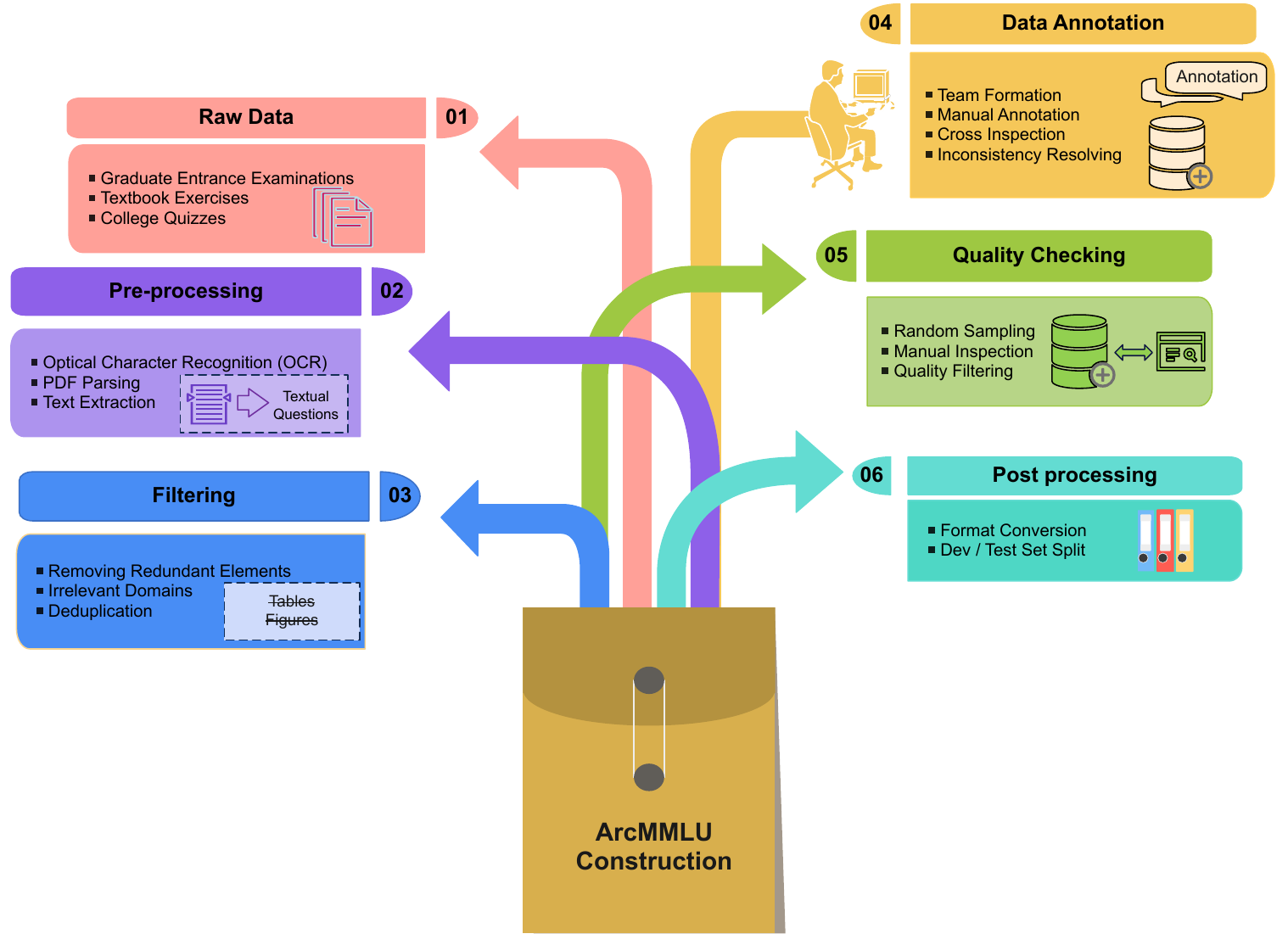}
\caption{The overall construction workflow of ArcMMLU.}
\label{fig:workflow}
\end{figure*}

Large language models (LLMs) have made remarkable advancements in recent years, pushing the boundaries of AI to unprecedented heights \cite{brown2020language, hoffmann2022training, touvron2023llama, touvron2023llama2, openai2023gpt4, li2023batgpt, zhang2023arcgpt}. Consequently, evaluating LLM capabilities has become increasingly complex and challenging. Traditional benchmarks like GLUE \cite{wang2019glue} and SuperGLUE \cite{wang2020superglue} have become insufficient for LLMs evaluation, as they lack the nuance and difficulty to effectively differentiate the advanced capabilities of newly developed models.

To address this, comprehensive benchmarks such as MMLU \cite{hendrycks2021measuring} and CMMLU \cite{li2023cmmlu} have been introduced. These benchmarks cover a broad spectrum of subjects, ranging from mathematics to social science, becoming essential tools for assessing LLMs' knowledge and reasoning capabilities.

However, there exists a notable void in LLM evaluation within the Library \& Information Science (LIS) domain. The LIS domain possesses a distinct set of disciplinary knowledge that general benchmarks like MMLU and CMMLU cannot fully capture. Professionals in the LIS field, such as archivists and librarians, urgently require the assistance of LLMs to enhance their work efficiency. Yet, the development of LIS-domain-focused LLMs is hindered by the lack of comprehensive, domain-specific model evaluation tools. Such tools are essential for accurately accessing the effectiveness of LLMs in addressing the unique challenges and tasks faced by LIS professionals, such as complex information retrieval, data organization, and user interaction in archives\/library settings. Therefore, there is a compelling need for a specialized LIS domain benchmark to accurately and thoroughly assess models' capabilities in this specialized area.

In this study, we introduce ArcMMLU, a Chinese-oriented benchmark specifically designed for evaluating LLMs in the LIS domain. We put our focus on four key subdomains: Archival Science, Data Science, Library Science, and Information Science. To compile a comprehensive benchmark that can effectively reflect the nature of LIS and provide robust evaluation, we have collected over 6,000 high-quality single-choice questions following the same format of MMLU\/CMMLU. Also, meticulous cleaning, filtering, and inspection have been performed to ensure the data quality.

Based on the constructed ArcMMLU, we conduct an extensive evaluation of the mainstream LLMs, including GPT-4. The evaluation results reveal that while most mainstream LLMs achieve an average accuracy rate above 50\% on ArcMMLU, there remains a notable performance gap, suggesting substantial headroom for improvement.

To uncover insights for guiding future development, we conduct further analysis which includes: (1) exploring the effect of few-shot examples on model performance; (2) performing error analysis on the challenging questions where current models consistently underperform. From the results, we observe that some models' performance diminishes with the insertion of more few-shot examples, indicating potential test data leakage during their training phase. Additionally, our error analysis reveals a notable deficiency in LIS-domain-specific knowledge. This finding suggests the need for incorporating more LIS-related data in training to enhance LLMs' domain knowledge.

The contribution of this work can be summarized in three folds:
\begin{enumerate}
    \item We construct a comprehensive LIS-domain-focused LLM evaluation benchmark, ArcMMLU, which covers four key subdomains with over 6,000 high-quality real-world questions.
    \item We evaluate the mainstream LLMs on ArcMMLU and the results reflect the current models' capabilities and limitations across different subdomains.
    \item We conduct further analysis to identify critical areas for potential improvement, providing insights for future LLM development in LIS domain. 
\end{enumerate}

\section{Related Work}

Benchmark evaluations play an essential role in tracking AI advancements, particularly in the rapidly evolving field of LLM. In the LLM era, traditional NLP benchmarks \cite{wang-etal-2018-glue, xu-etal-2020-clue, wang2020superglue} are increasingly proving to be inadequate. On one hand, these benchmarks focus on assessing model performance in specific tasks or dimensions but do not comprehensively reflect a model's capabilities across diverse domains and tasks. On the other hand, the level of challenge posed by these benchmarks is often too simplistic for cutting-edge LLMs \cite{openai2023gpt4, touvron2023llama2}, failing to differentiate between various models.

To address the limitations of traditional benchmarks, LLM-oriented benchmarks have been introduced \cite{hendrycks2021measuring, chen2021evaluating, cobbe2021training, srivastava2023imitation, zellers2019hellaswag, lin2022truthfulqa}. These benchmarks are either more comprehensive or more challenging, making them robust platforms for evaluating advanced LLMs. For Chinese-oriented models, similar benchmarks such as C-Eval \cite{huang2023ceval} and CMMLU \cite{li2023cmmlu} have been developed. These benchmarks are designed to evaluate models across a broad spectrum of subjects, posing a high challenge for model knowledge. Moreover, other benchmarks focus on task-specific or domain-specific evaluation, such as LawBench \cite{fei2023lawbench} and PIXIU \cite{xie2023pixiu}, which assess models in specialized areas. 

Despite that various benchmarks \cite{zeng2023measuring, liu2023m3ke, zhong2023agieval} have been introduced, there exists a void in the LIS domain.
Such void stems from the uniqueness of the LIS domain. Unlike more general fields, LIS encompasses a distinct and relatively self-contained set of disciplinary knowledge, often not adequately covered by comprehensive benchmarks like CMMLU. ArcMMLU aims to bridge such gap by providing a tailored evaluation platform. It focuses on the specific needs and knowledge inherent to LIS, such as information retrieval and archival management. Based on ArcMMLU, LLMs that are better suited to the unique requirements of the LIS field can be developed.

\begin{figure*}[ht!]
\centering
\includegraphics[scale=0.9]{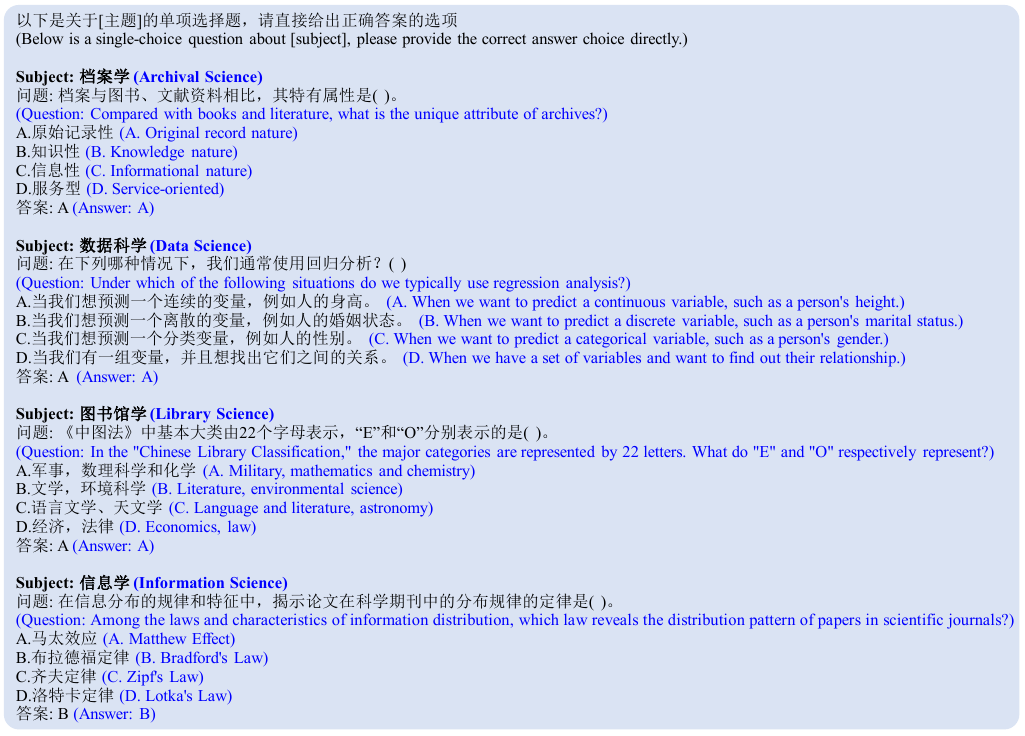}
\caption{Prompt examples from ArcMMLU.}
\label{fig:prompt-examples}
\end{figure*}

\section{ArcMMLU}

\paragraph{Task Overview}


ArcMMLU, our LIS-focused evaluation benchmark in Chinese, covers four key subdomains: Archival Science, Data Science, Library Science, and Information Science. Each of these subdomains has its distinct focus: Archival Science on the management and utilization of archival resources; Data Science on the processing and analysis of data; Library Science on the organization of library resources; and Information Science on information retrieval and management. 
We recognize that definitions of LIS disciplinary scope vary across different countries and regions and LIS encompasses more than these four areas. We plan to incorporate more subareas of LIS in future updates to provide a more comprehensive view of the field.


\paragraph{Construction}


To build a high-quality benchmark, we designed and implemented a meticulous construction process. The overall workflow of constructing ArcMMLU is presented in Figure~\ref{fig:workflow}. In the preparation phase of the project, we formed a team comprising 8 graduate students all with a background in LIS and 2 experienced professors as supervisors. The raw data was sourced from real-world questions of past postgraduate entrance exams, professional exams, course quizzes, and academic competitions to authentically reflect the characteristics of the LIS domain. Some original examples were stored in image format, from which we extracted textual content using OCR, followed by manual post-OCR result checking and quality filtering. During the filtering stage, we removed all samples containing images or tables that could not fit into text-based queries for LLMs. For the unlabeled samples in the raw data, we conducted manual labeling, with the supervisors resolving any inconsistencies among different annotators. The total labor cost of data collection, preprocessing, annotation, and inspection is around 300 hours. After quality filtering, we obtained a total of 6,210 samples. From these, we selected 20 representative samples as few-shot examples. The final statistics of ArcMMLU are detailed in Table~\ref{tab:stats}.

\begin{table}[ht]
\centering
\begin{tabular}{lrr}
\toprule
Knowledge Area        & Dev & Test \\ 
\midrule
Archive Science       & 5   & 2213 \\
Information Science   & 5   & 1674 \\
data Science          & 5   & 1499 \\
Library Science       & 5   & 804  \\
\midrule
Total                 & 20  & 6190 \\
\bottomrule
\end{tabular}
\caption{Statistics of the dev and test sets of ArcMMLU.}
\label{tab:stats}
\end{table}

\section{Experiments}

To gain a comprehensive understanding of existing open-source Chinese-oriented LLMs, our study evaluates models from six leading model families. By comparing and analyzing their results on the ArcMMLU benchmark, we can assess their knowledge and reasoning capabilities in the LIS domain and identify potential areas for improvement.

\subsection{Setup}

In ArcMMLU, all examples are single-choice questions with four options, where only one option is correct. For a given question, models need to select the label representing the correct answer from ‘A’, ‘B’, ‘C’, and ‘D’. We designed regular expressions to ensure that the model's output label is correctly recognized and extracted. To ensure fairness, if a model's output does not contain any meaningful label, we randomly sample a label from a uniform distribution as the response for that sample. Thus, a randomly initialized model would have a baseline performance of around 25\% on ArcMMLU.

We followed the prompt format of CMMLU. For each question, we prefix it with “以下是关于[主题]的单项选择题，请直接给出正确答案的选项 (Below is a single-choice question about [subject], please provide the correct answer choice directly)." In zero-shot evaluations, questions are immediately presented following the prompt. In few-shot evaluations, up to 5 examples with their corresponding answers are provided prior to the question. Prompt examples from ArcMMLU are shown in Figure~\ref{fig:prompt-examples}.

\begin{figure*}[h]
\centering
\includegraphics[width=1\textwidth]{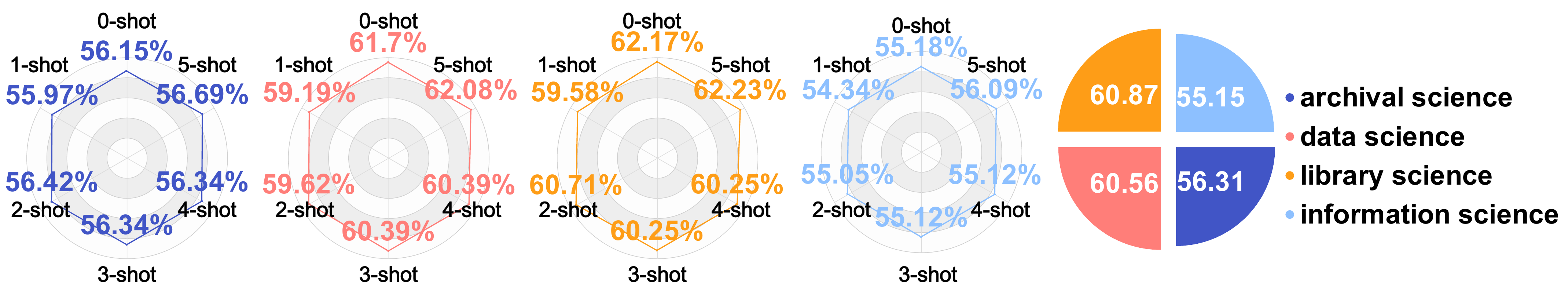}
\caption{Comparison of averaged accuracy across four subdomains (Archival, Data, Library, and Information Science) under different few-shot settings.}
\label{fig:by-domain}
\end{figure*}

\subsection{Models}

In our comprehensive evaluation, we test mainstream Chinese-oriented open-source Large Language Models (LLMs) using the ArcMMLU benchmark. We also include the leading multilingual models, ChatGPT and GPT-4, for broader comparative analysis. Specifically, these models belong to six distinct model families:

\paragraph{ChatGLM Series} Developed by Tsinghua University \cite{zeng2022glm}, the ChatGLM series are bilingual models in Chinese and English. They are based on the unique GLM pre-training strategy \cite{du2022glm} and are highly popular in the open-source community. In this study, we test both ChatGLM and ChatGLM2.

\paragraph{Qwen Series} Created by Alibaba Cloud \cite{bai2023qwen}, Qwen models have been pre-trained on a large corpus of 2.4 trillion tokens. We evaluate both the 7B and 14B versions to access their capabilities.

\paragraph{Baichuan Series} Developed by Baichuan Inc. \cite{yang2023baichuan}, these bilingual models are also trained on extensive high-quality corpora. For instance, Baichuan2 was pre-trained with 2.6 trillion tokens.

\paragraph{XVERSE Series} The XVERSE models, developed by Shenzhen Yuanxiang Technology and available in 7B and 13B sizes, have shown leading performance in various comprehensive benchmarks.

\paragraph{InternLM Series} InternLM, developed by SenseTime, comes in two sizes: 7B and 20B. We evaluated both to assess their performance on ArcMMLU.

\paragraph{ChatGPT and GPT-4} Developed by OpenAI, these globally leading LLMs are included to provide additional comparisons. By comparing existing open-source models with these, we aim to provide guiding insights for the future development of Chinese-oriented models.

\begin{table*}[htbp]
\centering
\setlength{\tabcolsep}{10pt}
\begin{tabular}{lccccc}
\toprule
Model        & Archival & Data  & Library  & Information & Average  \\ 
\midrule

GPT4          & 68.41   & 81.99         & 79.51       & 70.40   & 75.08 \\
Qwen-14B      & 67.24   & 72.85         & 71.39       & 64.80   & 69.07 \\
Baichuan2-13B & 62.01   & 66.91         & 66.07       & 61.44   & 64.11 \\
XVERSE-7B     & 61.91   & 63.44         & 65.77       & 61.69   & 63.20 \\
XVERSE-13B    & 60.33   & 62.98         & 65.53       & 60.20   & 62.26 \\
Qwen-7B       & 58.74   & 65.31         & 64.70       & 59.20   & 61.99 \\
Baichuan2-7B  & 59.56   & 63.51         & 63.26       & 59.95   & 61.57 \\
ChatGPT       & 53.95   & 68.18         & 66.31       & 53.86   & 60.57 \\
Baichuan-13B  & 56.12   & 61.24         & 62.19       & 59.33   & 59.72 \\
InternLM-20B  & 54.77   & 61.04         & 61.77       & 56.84   & 58.60 \\
InternLM-7B   & 50.75   & 59.24         & 59.44       & 50.50   & 54.98 \\
ChatGLM2-6B   & 52.78   & 53.97         & 51.85       & 48.76   & 51.84 \\
Baichuan-7B   & 50.70   & 50.63         & 52.63       & 46.64   & 50.15 \\
ChatGLM-6B    & 39.99   & 40.23         & 41.88       & 36.57   & 39.67 \\

\bottomrule
\end{tabular}
\caption{Results of all models under a five-shot setting. All results are reported in accuracy (\%).}
\label{tab:5-shot-results}
\end{table*}

\begin{table*}[ht!]
\centering
\setlength{\tabcolsep}{10pt}
\begin{tabular}{lccccc}
\toprule
Model        & Archival & Data  & Library  & Information  & Average  \\ 
\midrule

GPT4          & 66.38   & 82.12         & 78.55       & 66.79   & 73.46 \\
Qwen-14B      & 66.65   & 71.51         & 71.21       & 63.06   & 68.11 \\
Baichuan2-13B & 61.59   & 65.58         & 66.49       & 61.57   & 63.81 \\
XVERSE-7B     & 61.82   & 65.64         & 66.67       & 59.58   & 63.43 \\
Qwen-7B       & 59.78   & 64.31         & 63.68       & 56.47   & 61.06 \\
ChatGPT       & 52.37   & 65.64         & 66.19       & 52.61   & 59.20 \\
InternLM-20B  & 53.82   & 61.37         & 63.14       & 55.72   & 58.51 \\ 
Baichuan2-7B  & 56.08   & 61.11         & 61.11       & 55.72   & 58.50 \\
Baichuan-13B  & 54.41   & 57.17         & 61.05       & 54.48   & 56.78 \\
XVERSE-13B    & 53.37   & 56.73         & 59.62       & 57.21   & 56.73 \\
InternLM-7B   & 51.88   & 59.97         & 61.29       & 51.37   & 56.13 \\
ChatGLM2-6B   & 52.06   & 56.97         & 53.72       & 49.38   & 52.93 \\
Baichuan-7B   & 49.21   & 46.36         & 48.69       & 45.27   & 47.38 \\
ChatGLM-6B    & 43.06   & 46.96         & 47.91       & 38.81   & 44.19 \\
													
\bottomrule
\end{tabular}
\caption{Results of all models under a zero-shot setting. All results are reported in accuracy (\%).}
\label{tab:0-shot-results}
\end{table*}

\subsection{Main Results}

Table~\ref{tab:5-shot-results} and Table~\ref{tab:0-shot-results} present the performance of all models under a five-shot setting and a zero-shot setting, respectively.

\paragraph{By Model}

Analyzing the results presented in Table \ref{tab:5-shot-results} and Table \ref{tab:0-shot-results}, we gain insights into the performance of various LLMs across different subdomains. In both the five-shot and zero-shot settings, GPT4 demonstrates exceptionally powerful performance, reaching 73.46\% in zero-shot and 75.08\% in five-shot, significantly outperforming the second-ranked Qwen-14B, which scores 68.11\% and 69.07\% in zero-shot and five-shot, respectively. Notably, GPT4's lead is particularly significant in the data science and library science subdomains. For example, under a zero-shot setting, it outperforms other models by at least 10.61\% and 7.34\% in these two subdomains, respectively. Unlike GPT-4, ChatGPT, though also developed by OpenAI, demonstrates moderate performance, roughly matching the open-source Qwen-7B.

Another observable trend is that larger models generally exhibit better performance, with XVERSE-7B being an exception. Despite having only 7B parameters, it ranks just behind the larger Qwen-14B and Baichuan2-13B in both zero-shot and five-shot settings, even surpassing its own 13B version.

When comparing the zero-shot and five-shot methods, we note a difference in performance. Most models show performance improvements with the introduction of five-shot examples. However, some models, like ChatGLM-6B (dropping from 44.19\% to 39.67\%) and ChatGLM2-6B (from 52.93\% to 51.84\%), exhibit a decline in performance. In Section~\ref{sec:analysis}, we delve further into the analysis of these models that exhibit unexpected performance trends in the five-shot setting, exploring potential factors influencing these outcomes.

\begin{figure*}[h!] 
\centering
\includegraphics[width=0.9\textwidth]{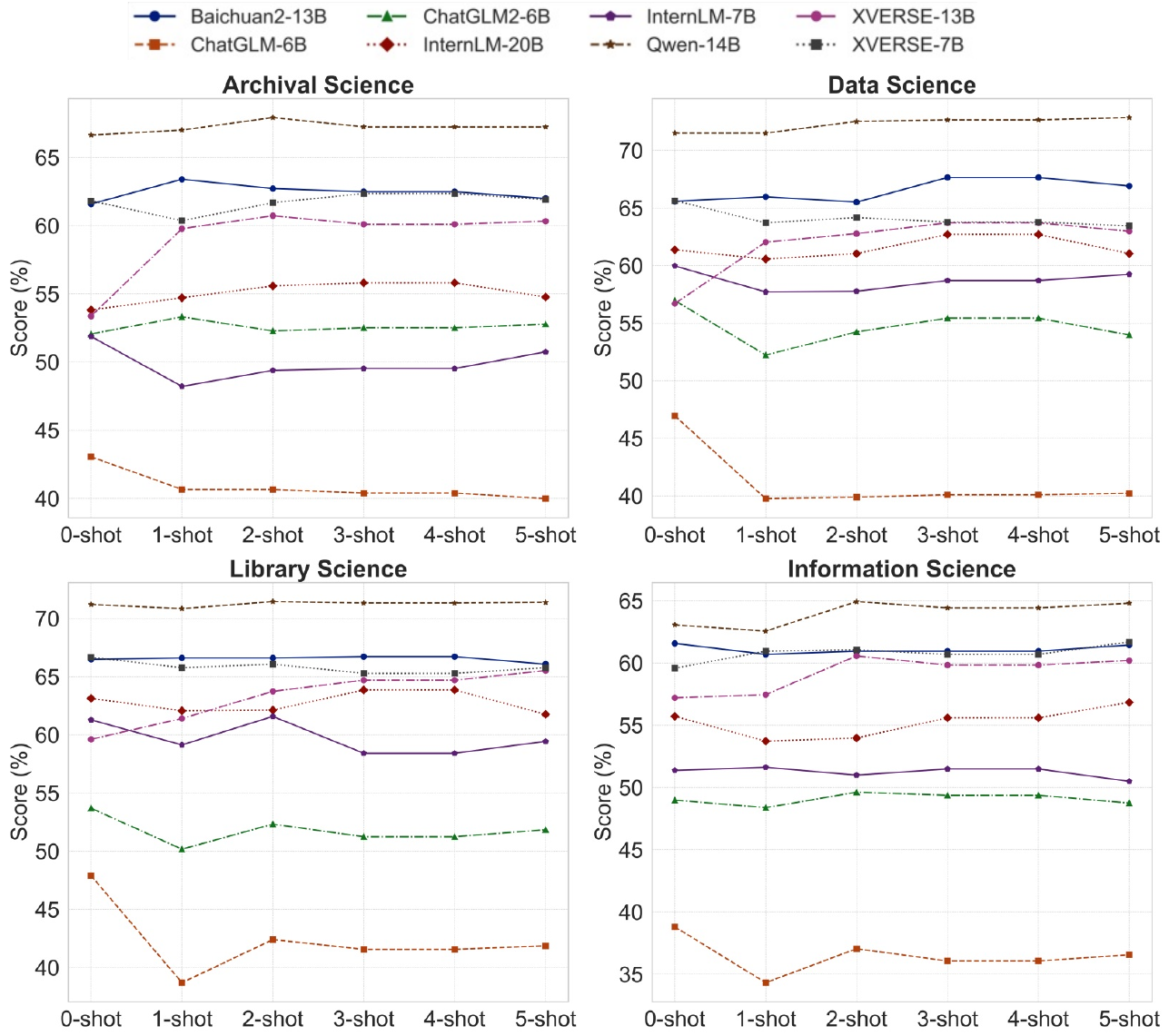}
\caption{The four knowledge domain's average performance of every LLMs from 0-shot to 5-shot}
\label{fig:few-shot-impact}
\end{figure*}

\paragraph{By subject}

Figure~\ref{fig:by-domain} shows the averaged accuracy of all models across four subdomains. We observe that models generally perform better in library science and data science, with averaged accuracies reaching 60.87\% and 60.56\%, respectively. In contrast, performance in archival science and information science is notably lower, at 56.31\% and 55.15\%, respectively. We speculate that this phenomenon may be caused by the interdisciplinary nature of data science and library science, which extensively overlap with other fields such as computer science and management. Hence, models are likely more exposed to information from these domains during the pre-training phase, resulting in stronger performance in these areas. On the other hand, archival science and information science have more self-contained disciplinary knowledge, and if models lack sufficient exposure to data distributions closely aligned to these subdomains during pre-training, it could lead to a deficiency in their capabilities. Based on this observation, future development of LIS-domain-specialized models should consider incorporating more archival and information science-related data in the pre-training phase to specifically enhance model performance in these areas.

Comparing the results under different few-shot settings, from 1-shot to 5-shot, all subdomains exhibit a trend where more few-shot examples lead to better performance. However, when contrasting zero-shot with few-shot results, the impact of introducing few-shot examples on performance varies across different subdomains. For instance, archival science surpasses its zero-shot result at 2-shot, but the other three subdomains only exceed their zero-shot performances at 5-shot. This suggests that the effectiveness of few-shot learning differs among subdomains, providing insights for tailored training approaches.

\section{Analysis}
\label{sec:analysis}

To further explore the behavior of LLMs in the LIS domain, we conduct a detailed analysis. Specifically, our analysis centers around four key aspects: (1) the usefulness of few-shot examples; (2) error analysis; (3) comparison with the leading model, GPT-4; and (4) potential directions for improvement.

\paragraph{Usefulness of Few-Shot Examples}
As mentioned in the previous section, we observed an unexpected phenomenon where some models' performance actually declines in the five-shot setting compared to zero-shot. To investigate the underlying reasons, we plot more fine-grained few-shot results from 0-shot to 5-shot, as shown in Figure~\ref{fig:few-shot-impact}. It is evident that most models either maintain a stable performance or show a gradual upward trend with more few-shot examples employed. XVERSE-13B, for example, shows significant performance gains in all four subdomains after the introduction of few-shot examples, demonstrating strong in-context generalization abilities. In contrast, some models, such as ChatGLM-6B, shows a performance decline in all four subdomains with few-shot examples introduced. This phenomenon can be partially explained by two key factors: (1) the fundamental capabilities of the model being inadequately robust, leading to weak in-context generalization, thus the introduction of few-shot examples actually disrupts the model’s context understanding, resulting in a performance drop; (2) potential test data leakage during the pre-training phase. Recent study \cite{wei2023skywork} reveals the risk that some existing LLMs have been exposed to data with a similar distribution to test data during pre-training, leading to overfitting issues on test data. This results in better performance under a zero-shot setting and a decline under a relatively unfamiliar few-shot setting. This observation suggests the necessity of highlighting the prevention of potential data leakage in future model development. Additionally, apart from direct data source restrictions, data from closely related fields that overlap extensively with the target domain can also indirectly lead to data leakage. For example, similar to ChatGLM-6B, ChatGLM2-6B also shows a performance drop in the data science and library science subdomains, which have more interdisciplinary overlap, whereas there is no such drop in the more self-contained subdomains of archival science and information science.

\paragraph{Error Analysis \& Comparison with GPT-4}

\begin{figure}[htbp] 
\centering
\includegraphics[width=0.5\textwidth]{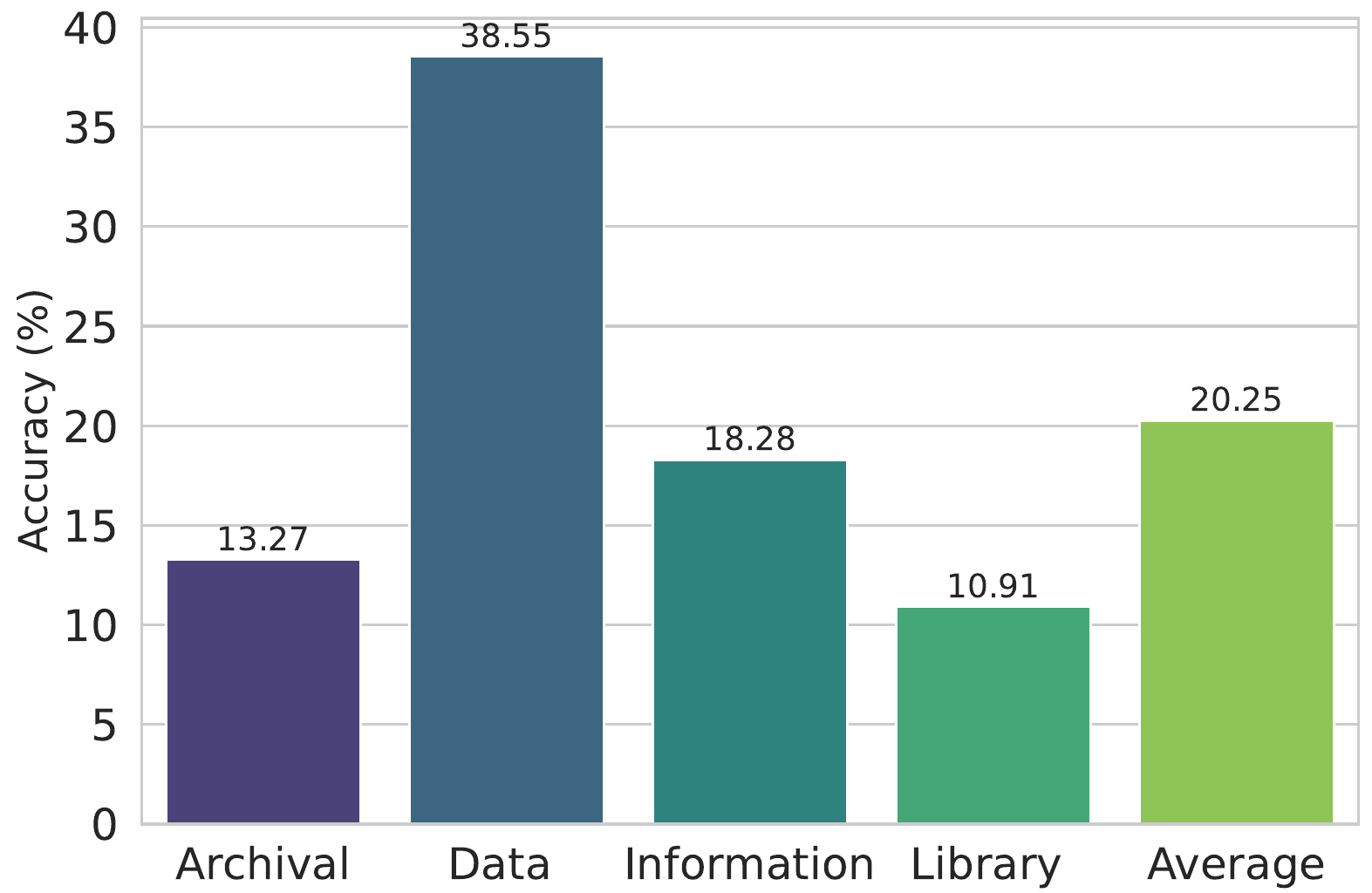}
\caption{Performance of GPT-4 on the ArcMMLU-Hard subset.}
\label{fig:gpt-4-on-hard}
\end{figure}

In LLM evaluations, error analysis is essential for highlighting their strengths and identifying their weaknesses. To understand the limitations of existing open-source LLMs, we extract questions that all open-source LLMs fail to answered correctly and compile them into a challenging test subset, named ArcMMLU-Hard. This subset implies a higher level of difficulty and challenge. We evaluate the performance of GPT-4 on this subset, and the results are detailed in Table~\ref{tab:gpt-4-on-hard} and Figure~\ref{fig:gpt-4-on-hard}. Notably, GPT-4 performs well on ArcMMLU-Hard, achieving an average accuracy of 20.25\%. Its performance is particularly high in Data Science, reaching 38.55\%. Upon manual inspection, we found that many of the Data Science questions in the ArcMMLU-Hard subset involve complex multi-step reasoning and calculations, posing a significant challenge for other models. GPT-4, however, manages these tasks more effectively, demonstrating its robust reasoning capabilities.

\begin{table}[htbp]
\centering
\begin{tabular}{lrrrr}
\toprule
Subdomain     & All & Corr. & Wrong & Acc. (\%) \\
\midrule
Archive       & 113 & 15      & 98    & 13.27         \\
Data          & 83  & 32      & 51    & 38.55         \\
Information   & 93  & 17      & 76    & 18.28         \\
Library       & 55  & 6       & 49    & 10.91         \\
\bottomrule
\end{tabular}
\caption{Statistics of the ArcMMLU-Hard subset and GPT-4's results.}
\label{tab:gpt-4-on-hard}
\end{table}

By manually checking GPT-4's correct answers in other subdomains of the ArcMMLU-Hard subset, we observe that GPT-4 not only possesses strong reasoning capabilities but also an impressive level of knowledge. For example, as shown in Appendix Figure~\ref{fig:gpt4-on-hard-correct}, GPT-4 correctly answers a question from Archival Science that tests a detailed regulation in Chinese archival management. This indicates that, although it is a multi-lingual general model, GPT-4 might have consumed a vast amount of Chinese data during its pre-training to achieve such powerful performance on a Chinese domain-specific evaluation.

Despite GPT-4's impressive performance, it is not infallible. Appendix Figure~\ref{fig:gpt4-on-hard-wrong} presents questions that both GPT-4 and other models answer incorrectly. These questions often involve (1) highly detailed and obscure domain-specific knowledge, and (2) easily confusable names and concepts. This observation highlights a critical limitation in the training of LLMs. Despite their extensive and diverse training data, these models struggle with questions requiring specific in-domain knowledge, which is not adequately injected during their training. This underscores the importance of incorporating a more comprehensive dataset that spans a broader spectrum of specialized knowledge domains. Addressing this limitation is essential for enhancing the models' ability to handle a wider range of challenging and specialized questions, thereby improving their overall applicability and reliability.

\section{Conclusion}



In this study, we introduce ArcMMLU, a comprehensive LLM evaluation benchmark specifically designed for the Chinese Library \& Information Science (LIS) domain. ArcMMLU covers four key subdomains: Archival Science, Data Science, Information Science, and Library Science, with over 6,000 high-quality questions obtained through meticulous collection, annotation, and inspection.
Our benchmark addresses the existing gap in LLM evaluation within the LIS domain. On one hand, it offers a robust platform to test existing models' knowledge and reasoning capabilities in the LIS domain. On the other hand, it provides insightful guidance for future development of LIS-domain-focused LLMs. Through our evaluation of mainstream LLMs on ArcMMLU, we identify significant areas for improvement. These include, but are not limited to, (1) preventing both direct and indirect test data leakage, and (2) incorporating more in-domain data during the pre-training stage to enhance models' domain knowledge and capabilities. Overall, ArcMMLU is a valuable addition to the current LLM evaluation landscape in the LIS domain and serves as a foundation for future development of LLMs within this specialized area.

\bibliographystyle{acl_natbib}

\clearpage

\appendix

\section{Additional Few-shot Results}

In this section, we present the experimental results under 1-shot to 4-shot settings, shown in Tables \ref{tab:results-one-shot} to \ref{tab:results-four-shot}.

\begin{table*}
\centering
\setlength{\tabcolsep}{10pt}
\begin{tabular}{llllll}
\toprule
Model        & Archival  & Data  & Library  & Information  & Average  \\ 
\midrule

Qwen-14B      & 67.01   & 71.51         & 70.85       & 62.56   & 67.98 \\
Baichuan2-13B & 63.40   & 65.98         & 66.61       & 60.71   & 64.17 \\
XVERSE-7B     & 60.37   & 63.71         & 65.77       & 60.95   & 62.70 \\
Qwen-7B       & 58.88   & 64.98         & 64.99       & 58.21   & 61.76 \\
Baichuan2-7B  & 58.79   & 62.51         & 62.31       & 59.83   & 60.86 \\
XVERSE-13B    & 59.78   & 62.04         & 61.41       & 57.46   & 60.17 \\
Baichuan-13B  & 56.76   & 60.44         & 61.41       & 58.33   & 59.23 \\
InternLM-20B  & 54.72   & 60.57         & 62.07       & 53.73   & 57.77 \\
InternLM-7B   & 48.22   & 57.71         & 59.14       & 51.62   & 54.17 \\
ChatGLM2-6B   & 53.32   & 52.23         & 50.18       & 48.38   & 51.03 \\
Baichuan-7B   & 49.66   & 48.83         & 51.49       & 46.02   & 49.00 \\
ChatGLM-6B    & 40.67   & 39.76         & 38.71       & 34.33   & 38.37 \\

\bottomrule
\end{tabular}
\caption{Results of all models under a one-shot setting. All results are reported in accuracy (\%).}
\label{tab:results-one-shot}
\end{table*}

\begin{table*}
\centering
\setlength{\tabcolsep}{10pt}
\begin{tabular}{llllll}
\toprule
Model        & Archival  & Data  & Library  & Information  & Average  \\ 
\midrule

Qwen-14B      & 67.92   & 72.52         & 71.45       & 64.93   & 69.20 \\
Baichuan2-13B & 62.72   & 65.51         & 66.61       & 60.95   & 63.95 \\
XVERSE-7B     & 61.68   & 64.18         & 66.07       & 61.07   & 63.25 \\
XVERSE-13B    & 60.73   & 62.78         & 63.74       & 60.57   & 61.95 \\
Baichuan2-7B  & 59.63   & 62.58         & 64.04       & 59.83   & 61.51 \\
Qwen-7B       & 58.97   & 64.78         & 64.87       & 57.21   & 61.46 \\
Baichuan-13B  & 55.94   & 60.51         & 59.98       & 56.84   & 58.31 \\
InternLM-20B  & 55.58   & 61.04         & 62.13       & 53.98   & 58.18 \\
InternLM-7B   & 49.39   & 57.77         & 61.59       & 51.01   & 54.94 \\
ChatGLM2-6B   & 52.28   & 54.24         & 52.33       & 49.63   & 52.12 \\
Baichuan-7B   & 51.56   & 49.63         & 53.35       & 47.51   & 50.51 \\
ChatGLM-6B    & 40.67   & 39.89         & 42.41       & 37.06   & 40.01 \\

\bottomrule
\end{tabular}
\caption{Results of all models under a two-shot setting. All results are reported in accuracy (\%).}
\label{tab:results-two-shot}
\end{table*}

\begin{table*}
\centering
\setlength{\tabcolsep}{10pt}
\begin{tabular}{llllll}
\toprule
Model        & Archival  & Data  & Library  & Information  & Average  \\ 
\midrule

Qwen-14B      & 67.24   & 72.65         & 71.33       & 64.43   & 68.91 \\
Baichuan2-13B & 62.49   & 67.65         & 66.73       & 60.95   & 64.45 \\
XVERSE-7B     & 62.36   & 63.78         & 65.29       & 60.72   & 63.03 \\
XVERSE-13B    & 60.14   & 63.71         & 64.71       & 59.83   & 62.08 \\
Qwen-7B       & 58.38   & 65.31         & 64.99       & 57.34   & 61.51 \\
Baichuan2-7B  & 59.15   & 63.64         & 62.72       & 60.23   & 61.43 \\
InternLM-20B  & 55.81   & 62.71         & 63.86       & 55.62   & 59.49 \\
Baichuan-13B  & 56.08   & 60.17         & 59.20       & 59.08   & 58.63 \\
InternLM-7B   & 49.53   & 58.71         & 58.42       & 51.49   & 54.54 \\
ChatGLM2-6B   & 52.51   & 55.44         & 51.25       & 49.38   & 52.14 \\
Baichuan-7B   & 51.97   & 50.77         & 52.93       & 46.39   & 50.51 \\
ChatGLM-6B    & 40.40   & 40.09         & 41.58       & 36.07   & 39.53 \\

\bottomrule
\end{tabular}
\caption{Results of all models under a three-shot setting. All results are reported in accuracy (\%).}
\label{tab:results-three-shot}
\end{table*}

\begin{table*}
\centering
\setlength{\tabcolsep}{10pt}
\begin{tabular}{llllll}
\toprule
Model        & Archival  & Data  & Library  & Information  & Average  \\ 
\midrule

Qwen-14B      & 67.24   & 72.65         & 71.33       & 64.43   & 68.91 \\
Baichuan2-13B & 62.49   & 67.65         & 66.73       & 60.95   & 64.45 \\
XVERSE-7B     & 62.36   & 63.78         & 65.29       & 60.71   & 63.03 \\
XVERSE-13B    & 60.12   & 63.71         & 64.71       & 59.83   & 62.08 \\
Baichuan2-7B  & 59.15   & 63.64         & 62.72       & 60.22   & 61.43 \\
Qwen-7B       & 58.38   & 65.31         & 64.99       & 57.34   & 61.51 \\
InternLM-20B  & 55.81   & 62.71         & 63.86       & 55.63   & 59.49 \\
Baichuan-13B  & 56.08   & 60.17         & 59.24       & 59.08   & 58.63 \\
InternLM-7B   & 49.53   & 58.71         & 58.42       & 51.49   & 54.54 \\
ChatGLM2-6B   & 52.51   & 55.44         & 51.25       & 49.38   & 52.14 \\
Baichuan-7B   & 51.97   & 50.77         & 52.93       & 46.39   & 50.51 \\
ChatGLM-6B    & 40.43   & 40.09         & 41.58       & 36.07   & 39.53 \\

\bottomrule
\end{tabular}
\caption{Results of all models under a four-shot setting. All results are reported in accuracy (\%).}
\label{tab:results-four-shot}
\end{table*}



\section{Error Analysis Examples}
\begin{figure*}[h!]
\centering
\includegraphics[width=0.9\textwidth]{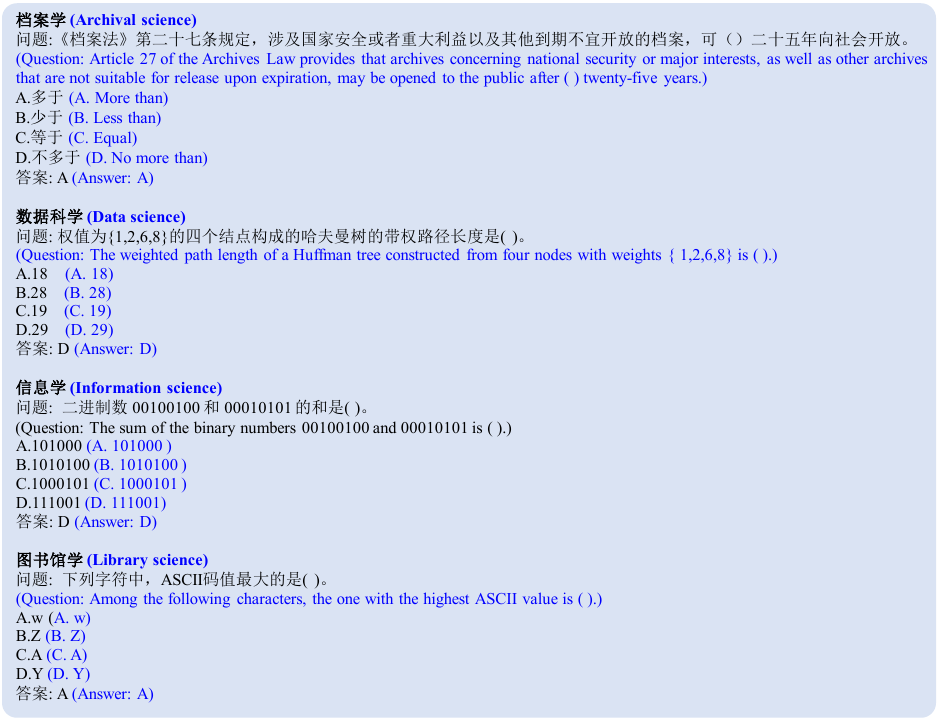}
\caption{Examples where GPT-4 provides correct answers while other open-source LLMs do not.}
\label{fig:gpt4-on-hard-correct}
\end{figure*}

\begin{figure*}[t!]
\centering
\includegraphics[width=0.9\textwidth]{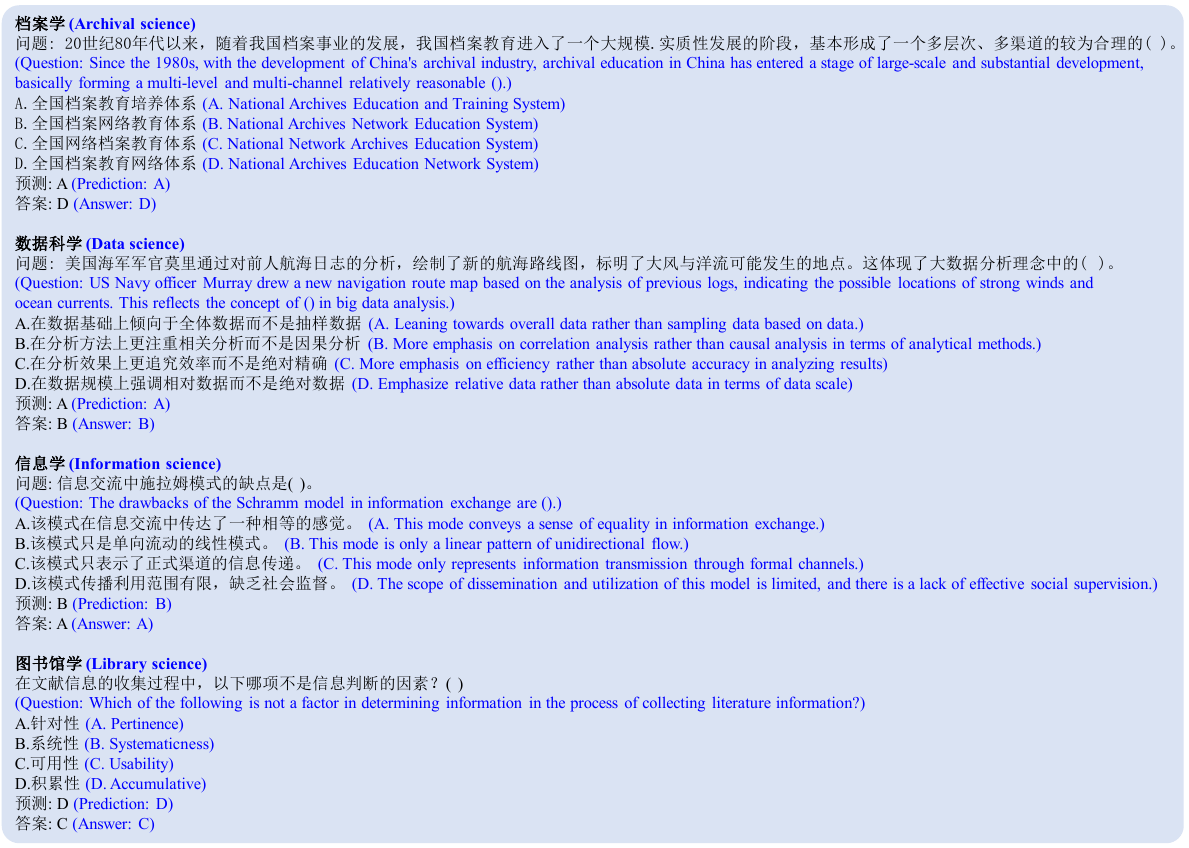}
\caption{Example where both GPT-4 and other open-source LLMs provide incorrect answers.}
\label{fig:gpt4-on-hard-wrong}
\end{figure*}

Figures~\ref{fig:gpt4-on-hard-correct} and~\ref{fig:gpt4-on-hard-wrong} present the samples from in our Error Analysis experiments.

\end{CJK*}  

\end{document}